\documentclass[10pt,twocolumn,letterpaper]{article}

\usepackage{listings}
\usepackage[pagenumbers]{cvpr} 

\usepackage{microtype}           
\usepackage{nicefrac}            
\usepackage{xspace}              
\usepackage{anyfontsize}
\usepackage{amsmath, amssymb, amsfonts, amsthm, mathtools,mathrsfs}
\usepackage{dsfont}              
\usepackage{fdsymbol}            
\usepackage[ruled,vlined]{algorithm2e}

\usepackage{booktabs}            
\usepackage{nicematrix}         
\usepackage{multirow}
\usepackage{makecell}
\usepackage{bigstrut}
\usepackage{enumitem}            
\setitemize{label=\textbullet, leftmargin=*, nolistsep}
\usepackage{graphicx}
\usepackage{adjustbox}           
\usepackage{float}               
\usepackage{wrapfig}             
\usepackage{subcaption}          
\expandafter\def\csname ver@subfig.sty\endcsname{}  
\usepackage{fontawesome5}        
\usepackage{pifont}              
\definecolor{cvprblue}{rgb}{0.21,0.49,0.74}
\usepackage[pagebackref,breaklinks,colorlinks,allcolors=cvprblue]{hyperref}
\usepackage[all]{hypcap}         
\usepackage{url}                 
\usepackage[normalem]{ulem}      
\usepackage{lipsum}              
\usepackage{rotating}            
\usepackage{stackengine}         
\usepackage{caption}             
\usepackage{listings}            
\usepackage{soul}                
\usepackage{gradient-text}       
\usepackage{pgfplots}            
\pgfplotsset{compat=newest}
\usepackage{pgfplotstable}       
\usepackage{tikz}                
\usepackage{svg}                 

\usepackage{wrapstuff}

\definecolor{demphcolor}{RGB}{125,125,125}             
\definecolor{cvprblue}{rgb}{0.21,0.49,0.74}
\usepackage[pagebackref,breaklinks,colorlinks,allcolors=cvprblue]{hyperref}
\usepackage{booktabs}      
\usepackage{multirow}      
\usepackage[table]{xcolor} 
\usepackage{graphicx}      
\usepackage{caption} 
\usepackage{tikz}
\usetikzlibrary{arrows.meta,positioning,fit,backgrounds,shapes.misc, calc}
\usepackage{pgfplots}
\pgfplotsset{compat=1.18}
\usepackage{enumitem}
\usepackage{pifont} 
\usepackage{adjustbox}
\usepackage{colortbl}
\usepackage{xcolor}
\usepackage{siunitx}

\definecolor{logoBlue}{HTML}{1A4E8A}
\definecolor{logoCyan}{HTML}{56BBCC}
\newcommand{\modelname}{\textbf{\textcolor{logoBlue}{3D-}\textcolor{logoCyan}{VCD}}}

\newcommand{\logoicon}{%
  \raisebox{-0.20\height}{\includegraphics[height=1.2em]{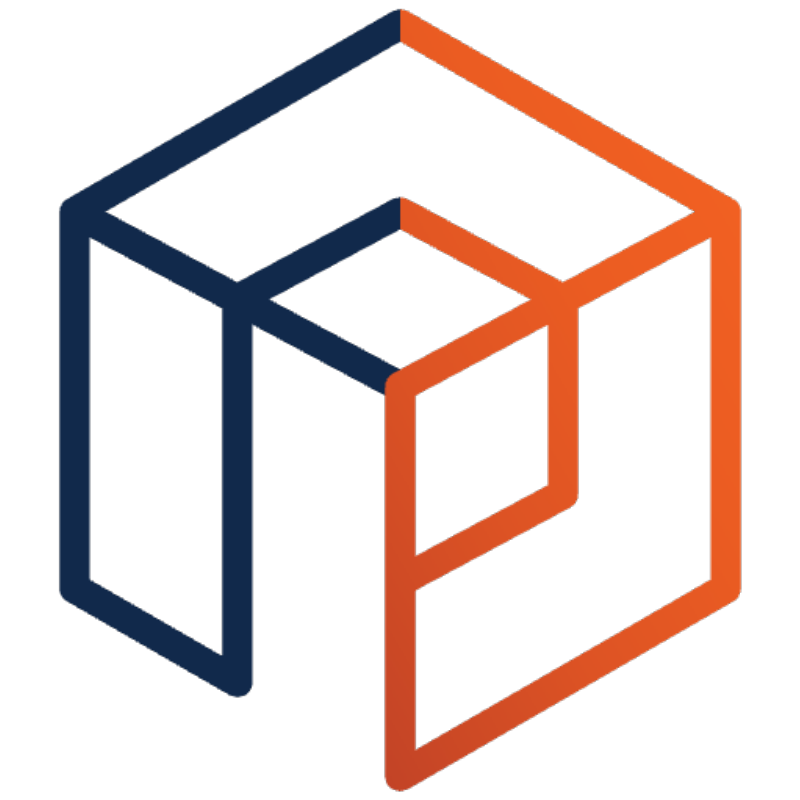}}%
}

\definecolor{IllinoisOrange}{HTML}{FF5F05}
\definecolor{IllinoisBlue}{HTML}{13294B}
\newcommand{\modelnamenc}{{{3D-}{VCD}}}

\def\paperID{38346} 
\def\confName{CVPR}
\def\confYear{2026}

\title{\modelname{}: Hallucination Mitigation in 3D-LLM Embodied Agents through \\ Visual Contrastive Decoding}

\author{Makanjuola Ogunleye$^{*}$\\
Virginia Tech\\
{\tt\small mogunleye@vt.edu}
\and
Eman Abdelrahman$^{*}$\\
Virginia Tech\\
{\tt\small emanh@vt.edu}
\and
Ismini Lourentzou\\
University of Illinois Urbana-Champaign \\
{\tt\small lourent2@illinois.edu}
}

\begin{document}
\maketitle
\begingroup
\renewcommand\thefootnote{}
\footnotetext{* Equal contribution.}
\endgroup
\begin{abstract}
Large multimodal models are increasingly used as the reasoning core of embodied agents operating in 3D environments, yet they remain prone to hallucinations that can produce unsafe and ungrounded decisions. Existing inference-time hallucination mitigation methods largely target 2D vision-language settings and do not transfer to embodied 3D reasoning, where failures arise from object presence, spatial layout, and geometric grounding rather than pixel-level inconsistencies. We introduce \modelname{}, the first inference-time visual contrastive decoding framework for hallucination mitigation in 3D embodied agents. 3D-VCD constructs a distorted 3D scene graph by applying semantic and geometric perturbations to object-centric representations, such as category substitutions and coordinate or extent corruption. By contrasting predictions under the original and distorted 3D contexts, our method suppresses tokens that are insensitive to grounded scene evidence and are therefore likely driven by language priors. We evaluate \modelnamenc{} on the 3D-POPE and HEAL benchmarks and show that it consistently improves grounded reasoning without any retraining, establishing inference-time contrastive decoding over structured 3D representations as an effective and practical route to more reliable embodied intelligence.

\noindent \logoicon~\href{https://plan-lab.github.io/3d-vcd}{\textcolor{IllinoisBlue}{PLAN Lab}~\textcolor{IllinoisOrange}{https://plan-lab.github.io/3d-vcd}}
\end{abstract}    
\begin{figure}[!t]
    \centering
    \includegraphics[width=0.99\linewidth]{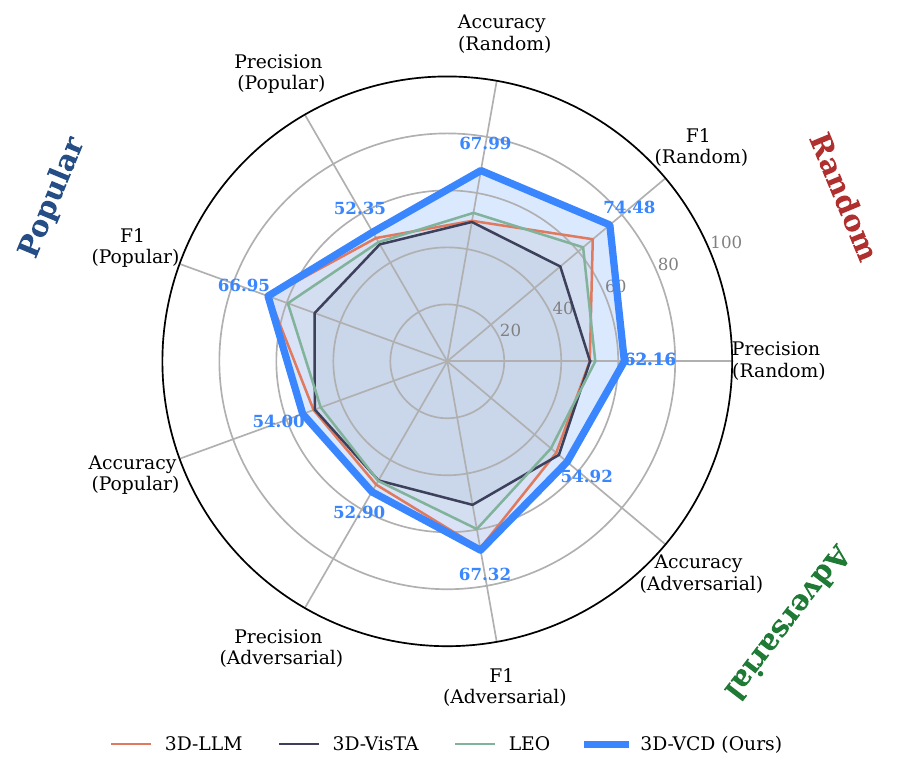}
    \vspace{-0.4cm}
    \caption{\textbf{Split-wise results on 3D-POPE benchmark.}
Each split (Random, Popular, Adversarial) shows Precision, F1, and Accuracy.
Our \modelnamenc{} training-free hallucination mitigation method consistently improves precision, F1, and accuracy while substantially reducing over-affirmation rates across all splits.}
    \label{fig:radar_chart_results}
\end{figure} 

\section{Introduction}
\label{sec:intro}
Multimodal Large Language Models (MLLMs) have demonstrated powerful reasoning and interaction capabilities across visual and textual modalities, sparking a surge of interest in their application to embodied AI \cite{liu2023visual, li2023blip2, alayrac2022flamingo, wang2024exploring, touvron2023llama,yu2025uncertainty}. By integrating MLLM backbones, embodied agents can follow natural-language instructions \cite{brohan2023can, zitkovich2023rt}, perform planning \cite{hu2024look, wu2025vsp}, answer spatial queries \cite{majumdar2024openeqa}, and operate in complex indoor environments \cite{driess2023palm, wang2023voyager, luo2025visual}. Recent progress in 3D-aware MLLMs has shown that integrating structured 3D information, such as scene graphs, point clouds, or volumetric features, significantly improves spatial grounding and object-centric understanding \cite{fu2024scene, liu2024uni3d, rana2023sayplan}.

However, despite these advances, embodied grounding hallucination remains a fundamental reliability bottleneck for embodied grounding.
Embodied agents frequently produce textually plausible responses that are inconsistent with the observed 3D scene, such as affirming the presence of nonexistent objects or misidentifying those that are present \cite{bai2024hallucination, huang2025survey, huang2024visual, rohrbach2018object}. These errors are especially problematic in embodied settings, where model outputs directly drive downstream action selection and physical interaction. As a result, hallucinations propagate into the control loop, leading to unsafe behaviors, degraded task performance, and fundamentally unreliable system behavior. Mitigating hallucinations is therefore essential for building embodied systems that act safely, reliably, and with trustworthy perceptual grounding.

Hallucinations have been documented and empirically studied extensively in recent analyses of 3D embodied models~\cite{chakraborty2025heal,chakraborty2025hallucination,huang2025architectural}, 
highlighting the severity of this issue: agents often default to distributional language priors when visual evidence is weak, ambiguous, or occluded, leading to high false-positive rates under scene–task inconsistencies. Yet, to the best of our knowledge, no hallucination mitigation strategy exists for 3D embodied agents.

A central challenge is generalization. Embodied agents must handle new tasks, novel object arrangements, and previously unseen 3D layouts. In this setting, training-based mitigation alone is inherently limited, because no dataset can exhaust the combinatorial diversity of real-world scenes. This makes inference-time mitigation especially appealing, as it can improve reliability without depending on how well the training data anticipates deployment conditions.

Existing inference-time hallucination mitigation strategies for multimodal understanding tasks, such as image captioning, visual question answering, segmentation, and other 2D settings~\cite{xie2024hallusionbench,yang2025madbench,leng2024mitigating,an2025mitigating,liu2025reducing, li2025counterfactual}, primarily treat hallucination as a 2D semantic inconsistency between text and pixels. These methods rely on perturbing pixel-space inputs and contrasting predictions under corrupted images. While effective for detecting semantic drift in static images, they do not transfer to embodied settings.
Embodied agents operate over multi-view, incomplete, and geometry-rich egocentric observations, where hallucinations arise from failures in 3D reasoning, object presence, occlusion, visibility, and spatial layout, rather than pixel-level inconsistencies. Pixel perturbations cannot create contradictory 3D evidence or probe whether a model’s predictions depend on spatial structure. 

To bridge this gap, we introduce \modelname{}, an inference-time Visual Contrastive Decoding (VCD) framework designed specifically for 3D embodied agents. Unlike 2D VCD, which perturbs pixel appearance, \modelnamenc{} leverages the agent’s structured 3D representations of the environment and produces controlled ``negative'' contexts via semantic and geometric distortions, such as object category, coordinate, and spatial extent manipulations. By contrasting logits under the original and distorted representations, \modelnamenc{} identifies and suppresses predictions insensitive to 3D evidence, enforcing visual–semantic consistency in real time.
To assess generality, we evaluate \modelnamenc{} on 3D-POPE \cite{yang20253d} and HEAL~\cite{chakraborty2025heal} embodied 3D grounding benchmarks, 
which probe object hallucinations in 3D MLLMs by asking whether a queried object exists within a reconstructed scene. 

As shown in Figure~\ref{fig:radar_chart_results}, experiments show that our proposed training-free \modelnamenc{} approach consistently improves grounded prediction quality across all 3D-POPE splits, reducing over-affirmation (Yes-rate) by 10.9\%--24.7\% and improving accuracy by 8.1\%--35.8\% relative to 3D-LLM. These results demonstrate that inference-time 3D contrastive reasoning is a promising direction for more reliable embodied intelligence. Even under combined semantic and geometric distortions, \modelnamenc{} maintains strong 3D grounding and outperforms the baseline across corruption settings.

\noindent In summary, the contributions of this work are the following:
\begin{itemize}
    \item We introduce \modelname{},  the first \emph{training-free}, inference-time contrastive decoding framework for hallucination mitigation in 3D embodied agents.
    \item We propose a simple and effective 3D counterfactual grounding mechanism that constructs distorted scene graphs through semantic and geometric perturbations, and uses dual-context logit fusion to suppress predictions unsupported by the underlying 3D evidence.
    \item We demonstrate that \modelname{} improves grounded reasoning across embodied hallucination benchmarks, reducing over-affirmation on 3D-POPE and lowering hallucination on HEAL with minimal computational overhead.
\end{itemize}

\section{Related Work}

\noindent \textbf{MLLMs for Embodied AI and 3D Reasoning.}
Recent embodied agents integrate Multimodal LLMs (MLLMs) \cite{huang2025mllmford, li2025seeground} to achieve open-vocabulary reasoning, grounding, and object-centric understanding in 3D scenes. Models such as 3D-LLM \cite{hong20233d}, 3D-VisTA \cite{zhu20233dvista}, LEO \cite{huang2024leo}, and SceneVerse \cite{jia2024sceneverse} inject 3D spatial representations, including scene graphs, volumetric features, or neural descriptors, into language models to enable spatial QA and language-driven interaction. While effective at 3D grounding, these models remain vulnerable to hallucination \cite{rawte2023survey}, often defaulting to language priors when visual grounding is ambiguous. Crucially, these works do not include any inference-time hallucination mitigation, instead relying solely on training or architectural grounding.
\vspace{0.1cm}

\noindent \textbf{Object Hallucination in 3D MLLMs.}
As embodied MLLMs become more capable, their reliability becomes paramount. Hallucinations, the tendency of multimodal large language models to generate plausible but ungrounded outputs, are a well-documented failure in general MLLMs \cite{zhou2024analyzing, chakraborty2025hallucination, rawte2023survey}, 
but far less explored in 3D embodied environments. This issue is, however, amplified in embodied settings, where a single hallucinated object can derail a task \cite{peng2025understanding, chen2024ll3da, yang20253d}.
HEAL \cite{chakraborty2025heal} demonstrates that 3D agents hallucinate frequently under scene–task inconsistencies and tend to over-rely on language priors. While 3D-POPE~\cite{yang20253d} provides a diagnostic benchmark revealing failure patterns, mitigation strategies in embodied settings are underexplored, with prior work reducing hallucination only indirectly through finetuning on more accurate scene representations \cite{yang20253d}. However, embodied settings contain long-tail layouts, noisy and cluttered environments, and substantial distribution shifts, under which training-based approaches fail to generalize.
\vspace{0.1cm}

\begin{figure*}
    \centering
    \includegraphics[width=\linewidth]{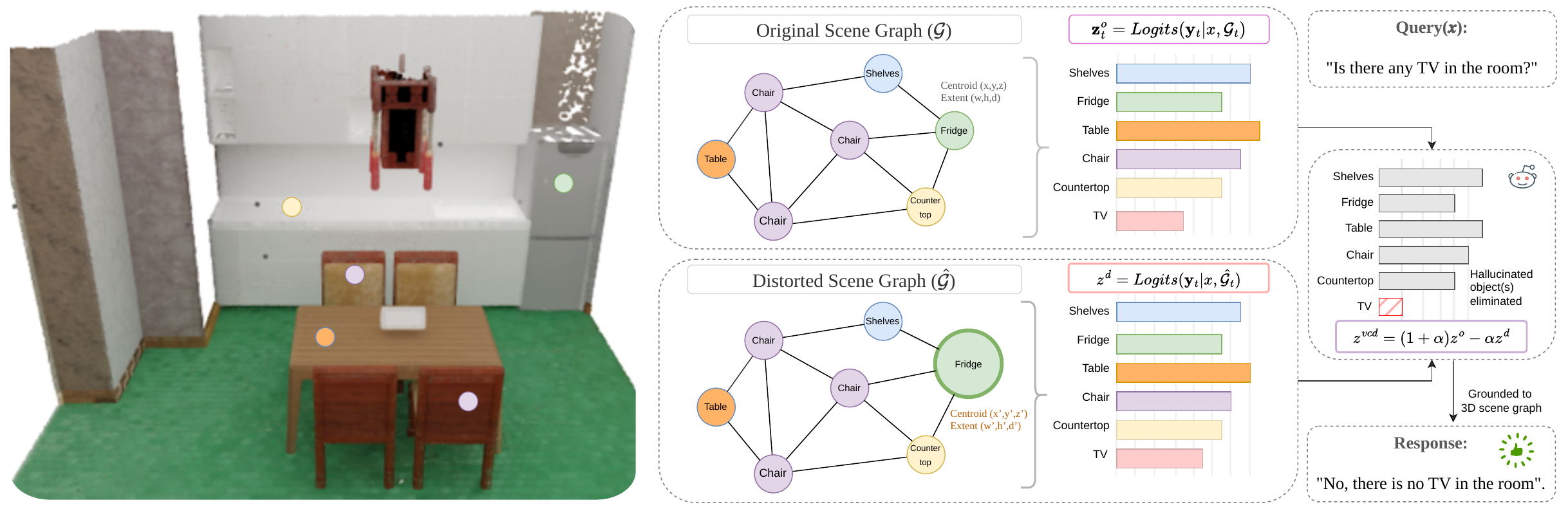}
    \vspace{-0.6cm}
    \caption{\textbf{\modelname{} Overview.}
    Given 3D environment observations, \modelname{} builds a structured 3D scene graph (${\mathcal{G}}$) encoding object categories, centroids, and extents, and injects controlled semantic and geometric perturbations to obtain a distorted version of the environment ($\hat{\mathcal{G}}$). The MLLM agent processes both contexts in parallel, given the textual query ($\mathbf{x}$). \modelnamenc{} contrastively fuses their logits to reveal and suppress hallucination-prone tokens. This training-free procedure enforces 3D-grounded reasoning at inference time.}
    \label{fig:overview}
\end{figure*} 

\noindent \textbf{Inference-time Hallucination Mitigation in MLLMs.}
Hallucination mitigation methods broadly include data-level, model-level, training-level, and inference-level approaches~\cite{bai2024hallucination}. Among them, inference-time interventions are appealing because they require no retraining. In 2D vision–language models, Visual Contrastive Decoding (VCD) \cite{leng2024mitigating, wang2024mitigating, park2025convis, lee2025retrieval} and its variants  \cite{park2025second,park2025convis,lee2025retrieval} contrast logits under original and visually distorted images to suppress language-only biases. These methods operate exclusively in 2D pixel space, relying on corruptions such as blurring, masking, or patch noise.
However, none of these methods apply to 3D embodied agents. In addition, pixel-level VCD is insufficient for embodied agents, as hallucination stems from errors in 3D structure and spatial reasoning.
To the best of our knowledge, our work is the first to introduce a training-free contrastive decoding framework for suppressing hallucinations in object-centric 3D embodied MLLMs.

\section{Method}
\label{sec:method}
We consider an embodied 3D reasoning setting in which an MLLM agent must answer natural-language queries conditioned on a structured perception of the surrounding 3D environment, that the agent observes at timestep  $t$, encompassing
multi-view RGB/D or point cloud observations. 
Given a natural-language query or referential question $\mathbf{x}_t$ and a structured 3D scene representation $\mathcal{G}_t$, the MLLM agent produces next-token logits $\mathbf{z}_t\!=\!f_{\theta}(\mathbf{x}_t, \mathcal{G}_t)$ and generates a textual response autoregressively $\mathbf{y}_{t,k}\!=\!\text{softmax}(\mathbf{z}_{t,k})$, $k=1, \dots, T$. 

However, 3D-MLLMs frequently over-rely on linguistic priors, assigning high probabilities to objects that are textually plausible but absent from the underlying 3D environment. This leads to hallucinations, \ie, responses that contradict the true scene, such as affirming nonexistent objects, misidentifying present ones, or defaulting to distributional language biases rather than 3D evidence. 

Hallucination can arise from weak perceptual grounding, distributional language biases, or spurious correlations learned during pretraining. In this work, we characterize hallucination based on the model's behavior under perturbed perceptual inputs. Specifically, hallucinated tokens are those whose predicted probabilities are not suppressed, and may increase, when the underlying 3D perceptual state is corrupted, \ie, when comparing $f_{\theta}(\mathbf{x}_t, \mathcal{G}_t)$ and $f_{\theta}(\mathbf{x}_t, \hat{\mathcal{G}}_t)$. This indicates that such predictions are not supported by grounded visual evidence but are instead driven by language priors.
Our goal in this work is to reduce such hallucinations directly during inference \emph{without} modifying the underlying MLLM architecture or retraining the backbone.

\subsection{3D Visual Contrastive Decoding (\modelnamenc{})}
\label{sec:3d-vcd-method}
To counteract this failure mode, we propose \modelnamenc{}, a training-free inference framework that mitigates 3D hallucinations by exploiting a structured, object-centric representation of the scene. As shown in Figure \ref{fig:overview}, \modelnamenc{} constructs a 3D scene graph that explicitly encodes the semantic and geometric attributes of each object, and introduces a novel family of graph-space distortions that perturb semantic labels, spatial geometry, and relational structure in a controlled, interpretable way. By evaluating the model under both the original and distorted graphs, \modelnamenc{} detects MLLM response tokens that are insensitive to 3D inconsistencies and contrastively suppresses them during decoding.
To the best of our knowledge, this is the first framework to leverage graph-level perturbations for 3D hallucination mitigation, enabling fine-grained reasoning about object presence while requiring no model retraining or architectural changes.

More formally, our method operates on a structured 3D object-centric scene graph at time step $t$, denoted as $\mathcal{G}_t$, defined as:
\begin{equation}
    \mathcal{G}_t = \{\, o_i = (c_i, a_i)\,\}_{i=1}^{N_t},
    \label{eq:scenegraph}
\end{equation}
where $N_t$ denotes the number of object nodes. Each node $o_i \in \mathcal{G}_t$ is defined by a semantic category $c_i$ and a set of structured attributes $a_i$ capturing object properties.
The structure of $a_i$ is benchmark-dependent. In 3D-POPE, $a_i$ consists of continuous geometric attributes such as 3D centroid and spatial extent. In contrast, HEAL represents $a_i$ using discrete state and relational attributes, including object states and explicit relations between object instances.

One instantiation of $\mathcal{G}_t$ used in 3D-POPE is shown below, where each object is denoted by a unique identifier:

\fbox{
\begin{minipage}{0.9\linewidth}
{\footnotesize
\begin{tabular}{@{}p{\linewidth}@{}}
\texttt{scene\_01: \{} \\
\texttt{\textbf{<obj$_1$>}: \{ category: "chair", } \\ 
\texttt{centroid:[$x_1$, $y_1$, $z_1$], extent:[$w_1$, $h_1$, $d_1$] \}}\\
...\\
\texttt{\textbf{<obj$_n$>}: \{ category: "table", } \\
\texttt{centroid:[$x_n$, $y_n$, $z_n$], extent:[$w_n$, $h_n$, $d_n$] \}}\\
\texttt{\}}\\
{\ttfamily \textbf{Query}: <refer\_expression> Are there any chairs in the room? <refer\_expression>}
\end{tabular}
}
\end{minipage}
}
\par \smallskip
This graph provides an explicit, disentangled representation of the environment that isolates the semantic and geometric attributes most relevant to hallucination in grounded reasoning.
Operating on this structured representation allows \modelnamenc{} to perturb object semantics, geometry, and relations in a controlled manner, producing corrupted graphs that remain syntactically valid for the MLLM. We consider two lightweight semantic and geometric perturbations:

\vspace{0.1cm}
\noindent \textbf{Graph-space distortions on 3D-POPE.} 
To probe the model's reliance on grounded scene information, \modelnamenc{} considers two lightweight semantic and geometric perturbations, implemented via a distortion operator $\mathcal{D}$ that constructs a corrupted scene representation $\hat{\mathcal{G}}_t = \mathcal{D}(\mathcal{G}_t)$, where $\mathcal{D}$ perturbs object-level attributes while preserving the structural schema of $\mathcal{G}_t$ required by the MLLM. Specifically, $\mathcal{D}$ modifies components of $a_i$, including semantic labels and geometric properties.

\begin{itemize}
    \item \textbf{Semantic perturbation:}
    \(
        \hat{c}_i \sim \mathrm{Shuffle}(c_i),
    \)
    replacing object categories with incorrect labels to contradict semantic evidence, forcing the model to rely on actual scene evidence rather than language priors.
    
    \item \textbf{Geometric perturbation:} 
    \(
        {\hat{a}_i = a_i + \epsilon,}
    \)
    {where the attribute vector is instantiated as $a_i = (p_i, s_i)$, with $p_i$ denoting the 3D centroid and $s_i$ denoting the spatial extent.} \modelnamenc{} applies noise to 3D centroids and extents to disrupt spatial grounding, where $\epsilon_p$ and $\epsilon_s$ $\sim \mathcal{N}(0, \sigma^2 I)$ denote small zero-mean Gaussian perturbations applied to object positions and extents, respectively. These distortions disrupt the physical configuration of the scene and test whether predictions change accordingly.
\end{itemize}
\vspace{0.1cm}

\noindent  \textbf{{Applying Distortions to the HEAL Benchmark.}} {HEAL introduces \emph{implicit} distortions through adversarial task formulations. In particular, HEAL provides paired task contexts consisting of a clean prompt and a corresponding adversarial variant (\eg, distractor injection, synonym substitution, or scene--task contradictions), which induce inconsistencies between language and the underlying 3D environment.
To adapt \modelnamenc{} to this setting, we reinterpret these adversarial prompts as distorted contexts without explicitly modifying the scene representation, where the scene representation $\mathcal{G}_t$ remains fixed, and the distortion is instead induced through the task specification. This formulation preserves the core principle of \modelnamenc{}: tokens that remain invariant across consistent and inconsistent contexts are indicative of hallucination driven by linguistic priors. By applying the contrastive fusion in ~\cref{eq:vcd}, we suppress such hallucination-prone predictions and promote responses that are sensitive to grounded scene information.
Importantly, this adaptation demonstrates that \modelnamenc{} is not restricted to explicit graph perturbations, but generalizes to broader forms of semantic and task-level inconsistencies. This enables a unified inference-time framework for hallucination mitigation across both geometry-centric settings (\eg, 3D-POPE) and higher-level embodied reasoning benchmarks (\eg, HEAL), without requiring additional training or architectural changes.}

\vspace{0.1cm}
\noindent \textbf{Dual-context inference.}
Given the original and corrupted 3D scene representations, the MLLM is evaluated under both contexts:
\begin{equation}
    \mathbf{z}_t^{o} = f_\theta(\mathbf{x}_t,\, \mathcal{G}_t), 
    \qquad
    \mathbf{z}_t^{d} = f_\theta(\mathbf{x}_t,\, \hat{\mathcal{G}}_t).
\end{equation}
Tokens whose logits are not suppressed under the perturbed context indicate that the prediction is not supported by the underlying 3D evidence and is likely driven by language priors. To suppress such hallucination-prone tokens, \modelnamenc{} contrastively fuses the outputs from the two contexts:
\begin{equation}
    \mathbf{z}_t^{\mathrm{vcd}} = (1+\alpha)\,\mathbf{z}_t^{(o)} \;-\; \alpha\, \mathbf{z}_t^{(d)},
    \label{eq:vcd}
\end{equation}
where $\alpha \ge 0$ controls the strength of the contrastive penalty. 
Intuitively, this penalizes predictions that remain highly probable under both graphs, while preserving or relatively favoring tokens that are supported by the true 3D scene. Autoregressive decoding then proceeds via
\begin{equation}
    \mathbf{y}_{t,k} = \mathrm{softmax}\!\left(\mathbf{z}_{t,k}^{\mathrm{vcd}}\right).
\end{equation}
This dual-stream contrastive formulation requires only one additional forward pass per query and introduces no additional trainable parameters, making  \modelnamenc{} sufficiently lightweight for embodied settings. {The proposed formulation is agnostic to the specific scene representation and applies to both geometry-centric and state-centric embodied reasoning settings.}

\vspace{0.1cm}
\noindent \textbf{Efficiency Considerations.}
To further minimize runtime overhead, we adopt two key optimizations that enable parallelized and memory-efficient inference:

 \noindent $\blackdiamond$ \textbf{Batched dual forward pass:} The original and distorted scene graphs are evaluated together in a single batched inference call to minimize redundant memory transfers and amortize model loading and tokenization costs.

 \noindent $\blackdiamond$ \textbf{Key--value caching:} For both the original and distorted graph contexts, transformer key-value (KV) states are cached after each decoding step, so that the model reuses all previously computed attention states rather than recomputing them from scratch, and only the new token is processed. This ensures that dual-context decoding incurs only a constant-factor overhead relative to standard autoregressive inference.
 
With these optimizations, \modelnamenc{} adds only a modest computational cost over baseline decoding, making it suitable for real-time or interactive embodied reasoning tasks. Empirically, in our experiments, baseline decoding requires approximately \textbf{2\,s} per query, while dual-context decoding in \modelnamenc{} takes only \textbf{2.5\,s} on average,  confirming that \modelnamenc{} adds negligible inference overhead with no retraining, making it practical for interactive embodied reasoning. 
\section{Experiments}
\label{sec:evaluation}
\begin{table*}[ht] 
\centering
\caption{\textbf{Results on the 3D-POPE benchmark.}
Across all three evaluation categories (Random, Popular, and Adversarial) \modelnamenc{}  achieves the highest precision, accuracy, and F1-score, surpassing prior 3D language models (3D-LLM, 3D-VisTA, and LEO). The substantial reduction in Yes-rate (e.g., \textbf{99.81\% → 75.15\%} in the Random set) alongside consistent gains in precision and accuracy demonstrates that VCD effectively mitigates over-affirmation bias and hallucination, yielding more balanced and reliable predictions in 3D reasoning.}
\vspace{-0.3cm}
\resizebox{1.0\linewidth}{!}
{
\begin{tabular}{@{}c@{}clc cccccc@{}}
\toprule
 & \textbf{3D-POPE} & \textbf{Model} & \textbf{Training-Free} & \textbf{Precision}$\uparrow$ & \textbf{Recall}$\uparrow$ & \textbf{F1 Score}$\uparrow$ & \textbf{Accuracy}$\uparrow$ & \textbf{Yes (\%)}$\downarrow$ \\
\midrule
 & \multirow{5}{*}{\textit{Random}} & 
    Random Baseline & - &  50.00 &  50.00 &  50.00 &  50.00 &  50.00 \\
    
    & & 3D-LLM~\citep{hong20233dllm} & \ding{56} &  50.03 &  99.88 &  66.67 &  50.07 &  99.81 \\
    & & 3D-VisTA~\citep{zhu20233dvista} & \ding{56} &  50.12 &  53.58 &  51.79 &  49.66 &  53.95 \\
    & & LEO~\citep{huang2024leo} & \ding{56} &  51.95 &  77.65 &  62.25 &  52.91 &  74.73 \\
   \cline{3-9}
\rowcolor{gray!20} \cellcolor{white} & \cellcolor{white}  & \textbf{\modelnamenc{} (Ours)} & \ding{52} &  \textbf{62.16} & 92.90 &  \textbf{74.48} &  \textbf{67.99} &  \textbf{75.15} \\
\cmidrule{2-9}
& \multirow{5}{*}{\textit{Popular}} & 
    Random Baseline & - &  50.00 & 50.00 &  50.00 &  50.00 &  50.00 \\
    & & 3D-LLM~\citep{hong20233dllm} & \ding{56} &  49.97 &   99.88 &  66.61 &  49.94 &  99.94 \\
    & & 3D-VisTA~\citep{zhu20233dvista} & \ding{56} &  47.40 &  51.88 &  49.54 &  49.49 &  52.30 \\
    & & LEO~\citep{huang2024leo} & \ding{56} &  48.30 &  77.65 &  59.55 &  47.27 &  80.38 \\
    \cline{3-9}
    \rowcolor{gray!20} \cellcolor{white} & \cellcolor{white}  &  \textbf{\modelnamenc{} (Ours)} & \ding{52} &  \textbf{52.35} &  92.86 &  \textbf{66.95} &  \textbf{54.00} &  \textbf{89.02} \\
\cmidrule{2-9}
& \multirow{5}{*}{\textit{Adversarial}} & 
    Random Baseline & - &  50.00 &  50.00 &  50.00 &  50.00 &  50.00 \\
    & & 3D-LLM~\citep{hong20233dllm} & \ding{56} &  49.97 &  99.88 &  66.61 &  49.94 &  99.94 \\
    & & 3D-VisTA~\citep{zhu20233dvista} & \ding{56} &  48.28 &  54.39 &  51.15 &  51.14 &  52.99 \\
    & & LEO~\citep{huang2024leo} & \ding{56} &  48.47 &  77.98 &  59.78 &  47.52 &  80.45 \\
    \cline{3-9}
\rowcolor{gray!20} \cellcolor{white} & \cellcolor{white}  & \textbf{\modelnamenc{} (Ours)} & \ding{52} &  \textbf{52.90} &  92.59 &  \textbf{67.32} &  \textbf{54.92} &  \textbf{87.82} \\
\bottomrule
\end{tabular}
}
\label{tab:3d_pope_main}
\end{table*}

\subsection{Experimental Setup}
\noindent \textbf{Benchmark.}
We evaluate the effectiveness of our proposed {\modelnamenc{}} method on the \textbf{3D–POPE}~\cite{yang20253d} and \textbf{HEAL}~\cite{chakraborty2025heal} benchmarks. These benchmarks are well-suited to test the robustness of our method as they probe 3D object and state hallucinations as well as object presence understanding through binary visual questions.

\noindent  \textbf{3D–POPE Benchmark.} Each instance in 3D–POPE contains a textual question (\eg, ``Is there a chair in the room?'') paired with a structured 3D scene graph representing the spatial and semantic layout of the environment.
The benchmark is divided into three subsets: \textit{random}, \textit{popular}, and \textit{adversarial}, 
targeting random sampling, frequency–biased objects, and hallucination–prone distractors, respectively. Following the 3D-POPE evaluation, we compute five quantitative metrics for each subset:
\textbf{Accuracy}, \textbf{Precision}, {Recall}, \textbf{F1-score}, and \textbf{Yes-rate}, which measures the proportion of affirmative responses and serves as a proxy for overconfidence or hallucination bias. A lower Yes-rates under adversarial subsets indicates stronger hallucination resistance. 

\noindent  \textbf{HEAL Benchmark.} The HEAL benchmark is a comprehensive hallucination probing set built on top of the Embodied Agent Interface Benchmark~\cite{li2024embodied}, capable of inducing hallucination rates up to $40\times$ higher than base prompts. The benchmark tests an agent's grounding by introducing five types of scene-task inconsistencies:
(1) \textit{Distractor Injection} (adding ungrounded objects to the task description),
(2) \textit{Object Removal} (removing task-critical objects from the scene representation),
(3) \textit{Scene-Object Synonymous} (using synonyms for objects between the scene and task),
(4) \textit{Scene-Task Contradiction} (providing a task that is impossible given the scene objects),
and (5) a \textit{Baseline} for comparison.
For this benchmark, we adopt the official HEAL metrics: \textbf{CHAIR} (Caption Hallucination Assessment with Image Relevance)~\cite{rohrbach2018object}. {CHAIR} is used to quantify the proportion of hallucinated objects or states compared to all mentioned entities as defined in \cref{eq:chair}. {CHAIR} is reported for both object hallucinations ($C_O$) and state hallucinations ($C_S$). Specifically, the metric is defined as
\begin{equation} \label{eq:chair}
\resizebox{0.9\linewidth}{!}{$
C_t = \frac{|\{\text{hallucinated } t\}|}{|\{\text{all } t \text{ mentioned}\}|},\ t \in \{\text{states}, \text{objects}\}
$}
\end{equation}

\noindent \textbf{Baselines.} We compare \modelnamenc{} against three representative 3D multimodal baselines finetuned on the 3D-GRAND dataset~\cite{yang20253d}:
(1) \textbf{3D–LLM}~\cite{hong20233dllm}, which integrates volumetric embeddings into large language models for open-vocabulary 3D reasoning;
(2) \textbf{3D–VisTA}~\cite{zhu20233dvista}, a vision–language model designed for embodied 3D understanding through spatial–semantic alignment; and
(3) \textbf{LEO}~\cite{huang2024leo}, that incorporates explicit 3D object grounding.
Unlike training-based baselines, for \modelnamenc{}, we use 3D-GRAND only to obtain object-level 3D context for scene graph construction, without any additional training.
\vspace{0.1cm}

\noindent \textbf{Implementation details.} 
For all 3D reasoning experiments, we build on the released 3D-LLM~\cite{hong20233dllm} model checkpoints from 3D-GRAND~\cite{yang20253d}, which is implemented on top of a LLaMA-style causal decoder. The model takes as input a text instruction and an object-centric scene description and autoregressively generates answers in natural language. We do not modify any model weights, as 3D-VCD operates purely at decoding time and is therefore model-agnostic. For experiments on HEAL, we additionally apply 3D-VCD to off-the-shelf instruction models (Llama-3-8B-Instruct and Qwen-14B-Instruct) by feeding them the 3D scene representations provided by the HEAL benchmark.  
For each iteration, we compute the baseline prediction (single pass) and the 3D-VCD prediction (two passes per decoding step). The extra overhead of computing is conceptually $2\times$ increase in forward-pass compute. However, with optimization strategies applied (Section \ref{sec:3d-vcd-method}), 3D-VCD increases end-to-end latency by $0.25\times$. All experiments were run on NVIDIA A40 GPUs with batch size 8.
We set the contrastive strength to $\alpha = 1.0$ by default and use greedy autoregressive decoding with temperature $T=1.0$. Additional implementation details are available in the Appendix.
\subsection{Results on 3D-POPE}
\label{sec:3dpope_results}
Table~\ref{tab:3d_pope_main} reports quantitative results of our proposed \textbf{\modelnamenc{}} on the {3D--POPE} benchmark. 
Across all three 3D-POPE evaluation settings, Random, Popular, and Adversarial, \modelnamenc{} achieves the highest overall performance, consistently outperforming existing 3D language models along every metric. Our method yields substantial gains in precision (\eg +10.0\% over the best baseline in the Random split), F1-score, and accuracy, while dramatically reducing the over-affirmation Yes-rate of 3D-LLM (\eg 99.81\%$\rightarrow$75.15\%). 

Under the \textit{Random} subset, \modelnamenc{} increases precision from 50.03\% to 62.16\% and accuracy from 50.07\% to 67.99\%. Moreover, on the \textit{Popular} and \textit{Adversarial} subsets, \modelnamenc{} achieves the highest F1 scores while maintaining recall above 92\% across all subsets.
These improvements demonstrate that contrastive evaluation of original and perturbed 3D contexts effectively suppresses hallucination-prone responses and produces more balanced, faithful predictions. Notably, \modelnamenc{} achieves these gains without any training, highlighting the practical value of inference-time hallucination mitigation.

\begin{table}[t!]
\centering
\caption{\textbf{Hallucination rates (CHAIR) on the HEAL probing set under the distraction injection prompt.} Applying 3D-VCD significantly reduces hallucinations and improves grounding for both Llama-3 and Qwen models to the factual objects and states present in the environment.} 
\vspace{-0.3cm}
\begin{tabular}{lcc}
\toprule
\textbf{Model} & \textbf{$C_O$ (\%)}$\downarrow$  & \textbf{$C_S$ (\%)}$\downarrow$  \\
\midrule
Llama-3-8B-Instruct-VCD   & \textbf{2.39} & 12.43 \\
Llama-3-8B-Instruct       & 2.58 & \textbf{9.49} \\
Qwen-14B-Instruct-VCD     & \textbf{3.55} & \textbf{5.00} \\
Qwen-14B-Instruct         & 4.13 & 16.45 \\
Gemma-2-9b-it             & 11.60 & 5.70 \\
DS-R1-Distil-LLaMA-8B     & 13.90 & 27.20 \\
\bottomrule
\end{tabular}
\label{tab:heal_distraction}
\vspace{-0.2cm}
\end{table}
\begin{figure*}[t]
    \centering
    \resizebox{\textwidth}{!}{%
    \begin{tikzpicture}
        \begin{axis}[
            ybar,
            bar width=15pt,
            ymin=0.5, ymax=0.8,                       
            nodes near coords,
            every node near coord/.append style={
                font=\small,
                rotate=90,
                xshift=12pt,
                yshift=-4pt,
                align=center,
                /pgf/number format/fixed,
                /pgf/number format/precision=3
            },
            ymajorgrids,
            grid style={dashed, gray!30},
            ylabel={F1},
            ylabel style={yshift=-4pt},
            symbolic x coords={
                Baseline,
                Struct-RelFlip,
                Low-SemSub-Geom,
                High-SemSub-Geom,
                Low-Geom,
                High-Geom,
                High-Sem-DropMod,
                Struct-Dist,
                High-SemSub,
                Low-SemSub,
                Struct-Sparse
            },
            xtick=data,
            x tick label style={
                rotate=25, anchor=east, align=right, font=\small
            },
            axis x line*=bottom,
            axis y line*=left,
            width=\textwidth,
            height=6.2cm,
            legend style={
                at={(0.5,1.10)},
                anchor=south,
                legend columns=-1,
                /tikz/every even column/.append style={column sep=0.6cm},
                nodes={scale=0.95, transform shape}
            },
            legend image code/.code={\draw[#1,rounded corners=1pt] (0cm,-0.08cm) rectangle (0.45cm,0.08cm);},
            x=2.15cm,  
            enlarge x limits=0.05,
        ]

        \addplot[draw=black, fill=violet!55] coordinates {
            (Baseline,      0.580)
            (Struct-RelFlip,        0.668)
            (Low-SemSub-Geom, 0.669)
            (High-SemSub-Geom, 0.668)
            (Low-Geom,     0.666)
            (High-Geom,     0.666)
            (High-Sem-DropMod, 0.668)
            (Struct-Dist,      0.669)
            (High-SemSub,      0.672)
            (Low-SemSub,      0.675)
            (Struct-Sparse,        0.672)
        };
        \addlegendentry{Adversarial}
        
        \addplot[draw=black, fill=orange!60] coordinates {
            (Baseline,      0.591)
            (Struct-RelFlip,        0.673)
            (Low-SemSub-Geom, 0.671)
            (High-SemSub-Geom, 0.676)
            (Low-Geom,     0.673)
            (High-Geom,     0.675)
            (High-Sem-DropMod, 0.674)
            (Struct-Dist,      0.673)
            (High-SemSub,      0.677)
            (Low-SemSub,      0.677)
            (Struct-Sparse,        0.680)
        };
        \addlegendentry{Popular}

        \addplot[draw=black, fill=teal!55] coordinates {
            (Baseline,      0.632)
            (Struct-RelFlip,        0.743)
            (Low-SemSub-Geom, 0.743)
            (High-SemSub-Geom, 0.743)
            (Low-Geom,     0.750)
            (High-Geom,     0.749)
            (High-Sem-DropMod, 0.749)
            (Struct-Dist,      0.749)
            (High-SemSub,      0.755)
            (Low-SemSub,      0.761)
            (Struct-Sparse,        0.770)
        };
        \addlegendentry{Random}

        \end{axis}
    \end{tikzpicture}%
    }
    \vspace{-0.8cm}
    \caption{\textbf{Ablation on distortion types for VCD in 3D-POPE (F1)}. $x$-axis shows distortion tags (ascending overall F1).}
    \label{fig:ablations_bars}
\end{figure*}
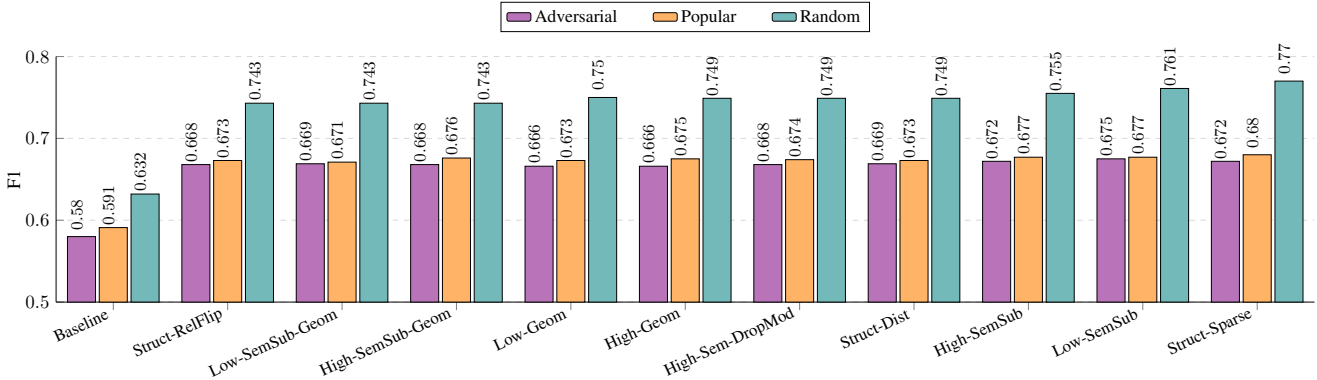%

\begin{figure*} [t]
    \centering
    \includegraphics[width=0.99\textwidth]{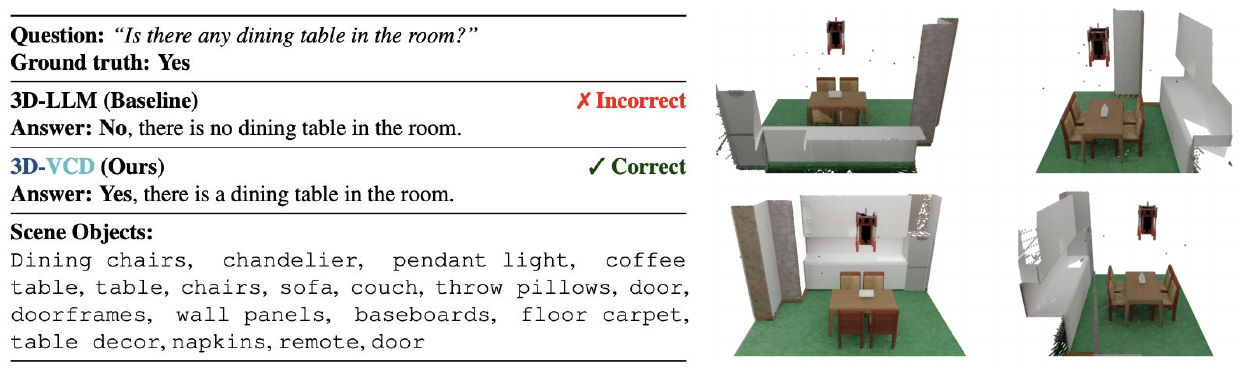}
    \vspace{-0.4cm}
    \caption{
    \textbf{Qualitative comparison on 3D-POPE.} The baseline incorrectly predicts the absence of a dining table, missing the true table object. In contrast, \modelnamenc{} correctly grounds and identifies the dining table by aligning contrastive decoding with the scene graph.
    }
    \label{fig:dining}
\end{figure*}

\subsection{Results on HEAL}
We assess \modelnamenc{} on the \textit{Distraction Injection} subset of the HEAL benchmark, a setting explicitly designed to induce scene–task inconsistencies by introducing misleading textual distractors. As shown in Table~\ref{tab:heal_distraction}, \modelnamenc{} substantially reduces both object and state hallucinations across models.

When applied to Qwen-14B-Instruct, 3D-VCD yields a dramatic improvement in state hallucination (\textit{CHAIR}–$C_S$), reducing its hallucination rate from 16.45\% to 5.0\% (a {3.3$\times$ reduction}). We also observe consistent gains in object hallucination (\textit{CHAIR}–$C_O$) for both Qwen-14B (4.13\% $\to$ 3.55\%) and Llama-3-8B (2.58\% $\to$ 2.39\%). These results demonstrate that \modelnamenc{} provides a strong grounding signal during decoding. By contrasting predictions under the true scene graph with those under a distorted context, our method effectively suppresses distractor-induced hallucinations and ensures generated plans remain faithful to the actual physical environment.\looseness-1

\subsection{Effect of different scene graph distortion types} 
\label{ablations}
We present a systematic study of how various {scene graph distortion types} affect the performance of our \modelnamenc{} framework. 
To isolate the contribution of different information cues, we apply one corruption type at a time, perturbing either {semantic}, {geometric}, or {structural} scene graph attributes:

\noindent \textbf{Semantic Corruptions.} 
We evaluate two levels of semantic perturbation:
(1) Semantic-Substitution (SemSub) replaces object category labels with near-synonyms or closely related categories 
(\eg, ``table'' $\rightarrow$ ``desk''
, probing the model’s robustness to lexical variation and category ambiguity. We further distinguish between two substitution levels: 
Low Substitution, where a small subset of object categories (approximately 10\%) are replaced, 
and High Substitution, where a larger portion (about 25\%) of categories are modified. 
(2) Semantic-DropModifier (SemDropMod) removes descriptive modifiers from compound object names 
(\eg, ``kitchen cabinet'' $\rightarrow$ ``cabinet''), probing the model’s sensitivity to reduced contextual specificity.
\begin{figure*} 
    \centering
    \includegraphics[width=0.99\textwidth]{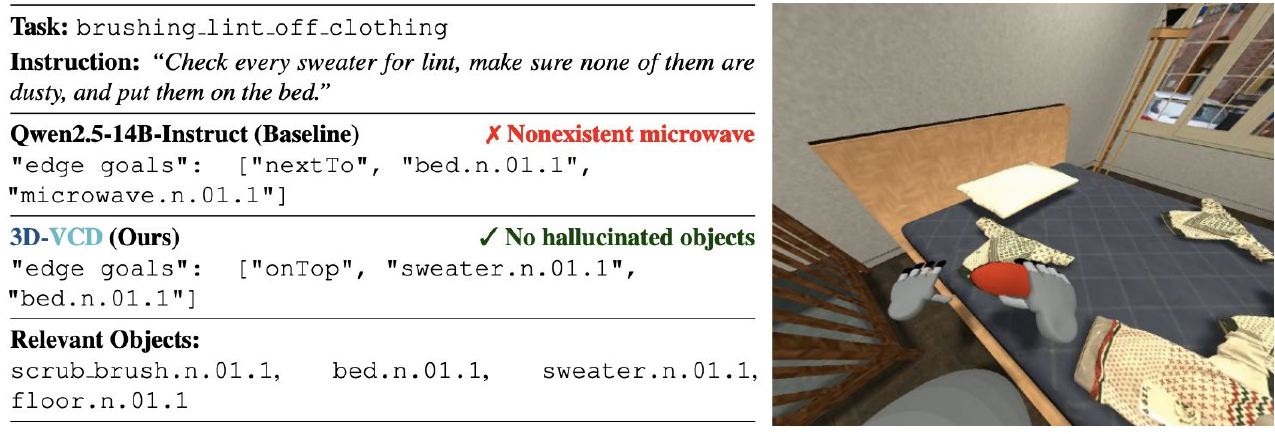}
    \vspace{-0.4cm}
   \caption{
\textbf{Qualitative comparison on HEAL benchmark.}
For the task \texttt{brushing\_lint\_off\_clothing}, the baseline Qwen-14B-Instruct model hallucinates a nonexistent 
\texttt{microwave.n.01\_1} in its symbolic goal prediction.
In contrast, \modelname{} produces clean symbolic goals with no hallucinated objects, correctly grounding all sweaters on the bed and removing dust-related states as required by the instruction.
The right panel shows the first-person view of the agent interacting with the sweaters in the scene.}
    \label{fig:lint}
\end{figure*}

\noindent \textbf{Geometric Corruptions.}
We apply controlled perturbations to object centroids and spatial extents by adding zero-mean Gaussian noise to the 3D coordinates. 
(1) Geometric-Low Noise (Low-Geom) introduces sensor-level jitter with a standard deviation of $\sigma=0.05\mathrm{m}$ for both centroid position and extents, simulating mild localization noise. 
(2) Geometric-High Noise (High-Geom) applies stronger Gaussian perturbations with $\sigma = 0.20\mathrm{m}$ for centroid position and extents, corresponding to large spatial displacements or poor reconstruction accuracy.
These settings evaluate the model’s tolerance to imprecise or corrupted 3D geometry.

\noindent \textbf{Structural Corruptions.}
We modify the underlying scene structure to assess the model’s reliance on relational and contextual cues.
Specifically, we perform (1) Object-Sparsification (Struct-Sparse) by randomly removing $\sim$20\% of scene objects. 
(2) Relation-Flipping (Struct-RelFlip) involves inverting 30\% of pairwise relations of spatial predicates, such as ``on top of'' or ``next to''. (3) Distractor-Injection (Struct-Dist) involves adding irrelevant objects or a few duplicate objects to simulate visual clutter and false detections.
These perturbations collectively test whether \modelnamenc{} depends on accurate relational structure rather than surface-level correlations.

\noindent \textbf{Mixed Corruptions.}
    Applies semantic and geometric perturbations jointly, mimicking realistic 
    embodied sensing noise where category uncertainty and spatial variance co-occur.

\noindent As shown in Figure \ref{fig:ablations_bars}, across all distortion types, \modelnamenc{} consistently outperforms the baseline model by a substantial margin. 
For example, in the \textit{Random} split, the baseline F1 score of 0.63 increases to {0.74–0.77} under different distortion variants. 
These gains demonstrate that controlled perturbations of the object-centric 3D scene representation reliably \emph{strengthen grounding and suppress hallucinations}. 
Moreover, the improvements remain stable across semantic, geometric, structural, and mixed distortions, indicating that \modelnamenc{} benefits broadly from contrasting consistent evidence against corrupted 3D cues.

Based on the balance between robustness, interpretability, and computational efficiency, we adopt the {mixed low semantic–geometric distortion} as our representative variant in subsequent experiments. 
By amplifying cues consistent with the true scene and down-weighting those insensitive to 3D changes, \modelnamenc{} promotes grounded reasoning over superficial correlations, thereby reducing hallucination and improving the reliability of 3D embodied agents.

\setlength{\columnsep}{7pt}
\begin{figure}[t!] 
\centering
\includegraphics[width=0.9\linewidth]{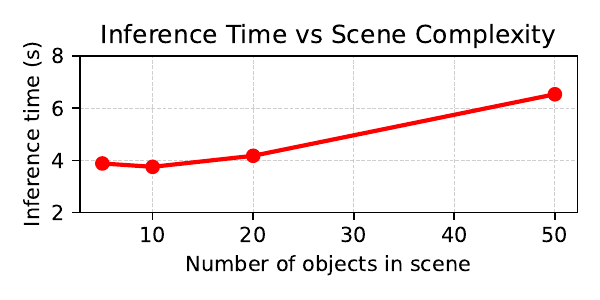}
\vspace{-0.6cm}
\caption{\textbf{\modelnamenc{} inference time} as a function of scene complexity (number of objects).}
\label{fig:runtime_scaling}
\vspace{-0.2cm}
\end{figure}

\subsection{Qualitative Results}
\label{qual_results} 
Figures \ref{fig:dining} and \ref{fig:lint}
present two qualitative examples illustrating how \modelnamenc{} improves factual grounding compared to the baseline. The first example (Figure \ref{fig:dining}) shows a 3D-POPE example in which the baseline underperceives the scene and incorrectly predicts that no dining table is present. Although the table is clearly represented in the object-centric scene description, the baseline answers ``No,'' revealing a failure to use grounded 3D evidence. In contrast, 3D-VCD correctly answers ``Yes'' by aligning decoding with the structured scene graph and reinforcing scene-consistent object evidence. 
Figure~\ref{fig:lint} shows a HEAL distractor-injection example in which the baseline hallucinates a nonexistent microwave, whereas 3D-VCD suppresses this spurious object and instead produces a clean grounded output that correctly focuses on the sweaters on the bed and the required state changes. This example demonstrates that 3D-VCD can effectively mitigate hallucinations by enforcing stronger consistency between language generation and scene-grounded evidence.

\section{Inference Efficiency}
We analyze inference time as a function of scene complexity (number of objects) and observe that our method scales smoothly with scene size, exhibiting only a modest increase in runtime as complexity grows. Importantly, 3D-VCD operates purely at decoding time without additional training or auxiliary models, making its overhead minimal and comparable to standard 3D-LLM inference. As shown in Figure \ref{fig:runtime_scaling} of our runtime analysis, inference time increases from approximately 3.8s for simple scenes to 6.7s for complex scenes with 50 objects, demonstrating that the method remains efficient and practical for real-world applications.
\section{Conclusion}
We introduce \textbf{\modelname{}}, an inference-time framework for mitigating hallucinations in 3D multimodal language models through a novel {contrastive decoding} strategy over structured scene representations. 
Unlike prior approaches that rely on retraining or pixel-space perturbations, \modelnamenc{} introduces a dual-scene contrastive process that operates on structured 3D scene graphs. This simple yet effective formulation substantially reduces over-affirmation bias while improving factual precision and grounding reliability across all 3D-POPE categories. 
Importantly, our method is training-free, architecture-agnostic, and incurs minimal computational overhead, making it readily applicable to existing 3D multimodal agents. 
Beyond its empirical gains, \modelnamenc{} highlights the potential of inference-time contrastive reasoning as an efficient mechanism for improving perception in embodied environments. 
Future work can explore extending this paradigm to temporal reasoning in dynamic 3D scenes.

\section*{Acknowledgments}
This research was partially supported by Google, the Google TPU Research Cloud (TRC) program, and the U.S. Defense Advanced Research Projects Agency (DARPA) under award HR001125C0303. The views and conclusions contained herein are those of the authors and should not be interpreted as necessarily representing the official policies, either expressed or implied, of Google, DARPA, or the U.S. Government. The U.S. Government is authorized to reproduce and distribute reprints for governmental purposes notwithstanding any copyright annotation therein.\\

{
    \small
    \bibliographystyle{ieeenat_fullname}
    \bibliography{bib2}
}

\clearpage
\appendix
\setcounter{page}{1}
\maketitlesupplementary

\section{Implementation Details}
We build on the released 3D-LLM~\cite{hong20233dllm} checkpoints from 3D-GRAND~\cite{yang20253d}, which adopt a LLaMA-style causal decoder that ingests object-centric scene graphs and autoregressively generates natural-language answers. We do not finetune or modify any model parameters; instead, \modelnamenc{} operates purely at inference time by contrastively manipulating token logits. Each scene is serialized into text using object category, centroid $(x,y,z)$, and extent $(w,h,d)$ attributes extracted from the dataset JSON. To construct the distorted negative context required for contrastive decoding, we inject zero-mean Gaussian noise into object centroids and extents, varying the standard deviation across $\sigma \in \{0.01, 0.05, 0.15, 0.45\}$ for sensitivity analysis. We prepend both the original and distorted scene tokens to the user query and perform dual decoding passes to obtain logit sequences $z_{\text{orig}}$ and $z_{\text{dist}}$, computing the final logits as $z_{\text{vcd}} = (1 + \alpha) z_{\text{orig}} - \alpha z_{\text{dist}}$ with $\alpha = 1.0$. We decode using greedy search with temperature $T=1.0$ and batch size 8. 
For HEAL experiments, we apply \modelnamenc{} to off-the-shelf language-only models (Llama-3-8B-Instruct and Qwen-14B-Instruct) by feeding them scene descriptions in the HEAL text format, requiring no spatial inference modules. Scene graphs are preprocessed once and cached for all runs. 

\begin{table*}[t]
\centering
\small
\caption{\textbf{Effect of Geometric Distortion Strength ($\varepsilon$)} on 3D-POPE Performance. Lower yes-rate indicates reduced hallucination.}
\vspace{-0.3cm}
\begin{tabular}{ccccccc}
\toprule
\textbf{3D-POPE} & \textbf{Geometric Distortions ($\boldsymbol{\varepsilon}$)} &
\textbf{Precision $\uparrow$ } & \textbf{Recall $\uparrow$} & \textbf{F1 $\uparrow$} & \textbf{Accuracy $\uparrow$} & \textbf{Yes \% $\downarrow$}  \\
\midrule

\multirow{5}{*}{\textit{Random}}
& \modelnamenc{} $\varepsilon{=}0.01$ & 59.20 & 98.02 & 73.81 & 63.86 & 86.05 \\
& \modelnamenc{} $\varepsilon{=}0.05$ & 61.89 & 95.17 & 75.00 & 67.65 & 78.41 \\
& \modelnamenc{} $\varepsilon{=}0.15$ & 61.68 & 95.28 & 74.88 & 67.39 & 78.82 \\
& \modelnamenc{} $\varepsilon{=}0.45$ & 59.21 & 98.25 & 73.89 & 63.79 & 86.55 \\
\midrule

\multirow{5}{*}{\textit{Popular}}
& \modelnamenc{} $\varepsilon{=}0.01$ & 51.54 & 97.66 & 67.47 & 52.45 & 95.69 \\
& \modelnamenc{} $\varepsilon{=}0.05$ & 52.35 & 94.09 & 67.27 & 53.34 & 91.61 \\
& \modelnamenc{} $\varepsilon{=}0.15$ & 52.48 & 94.44 & 67.47 & 53.46 & 91.96 \\
& \modelnamenc{} $\varepsilon{=}0.45$ & 51.67 & 97.55 & 67.56 & 52.63 & 95.46 \\
\midrule

\multirow{5}{*}{\textit{Adversarial}}
& \modelnamenc{} $\varepsilon{=}0.01$ & 50.37 & 97.26 & 66.37 & 51.32 & 95.35 \\
& \modelnamenc{} $\varepsilon{=}0.05$ & 51.50 & 94.39 & 66.64 & 53.30 & 90.57 \\
& \modelnamenc{} $\varepsilon{=}0.15$ & 51.60 & 94.04 & 66.64 & 53.28 & 90.42 \\
& \modelnamenc{} $\varepsilon{=}0.45$ & 50.43 & 97.25 & 66.42 & 51.39 & 95.33 \\
\bottomrule
\end{tabular}
\label{tab:geom_noise_ablation}
\end{table*}

\begin{table*}[t!]
\centering
\small
\caption{\textbf{Ablation on Semantic and Geometric Distortions} under varying contrastive strengths $\alpha$ on 3D-POPE.}

\vspace{-0.3cm}
\setlength{\tabcolsep}{4.5pt}
\renewcommand{\arraystretch}{1.2}
\begin{tabular}{llccccc}
\toprule
\textbf{3D-POPE} & \textbf{Method} &
\textbf{Precision $\uparrow$ } & \textbf{Recall $\uparrow$ } &
\textbf{F1 $\uparrow$ } & \textbf{Accuracy $\uparrow$ } & \textbf{Yes \% $\downarrow$} \\
\midrule
\multirow{5}{*}{\textit{Random}}

& Low-SemSub-Geom ($\alpha$=1.0)  & \textbf{62.16} & 92.90 & \textbf{74.48} & \textbf{67.99} & \textbf{75.15} \\
& Low-SemSub-Geom ($\alpha$=0.5)  & 59.19 & 96.22 & 73.29 & 64.39 & 82.56 \\
& High-SemSub-Geom ($\alpha$=1.0) & 59.52 & 95.65 & 73.38 & 64.58 & 81.99 \\
& High-SemSub-Geom ($\alpha$=0.5) & 59.42 & 95.95 & 73.39 & 64.56 & 82.26 \\

\midrule
\multirow{5}{*}{\textit{Popular}}
& Low-SemSub-Geom ($\alpha$=1.0)  & \textbf{52.35} & 92.86 & 66.95 & \textbf{54.00} & 89.02 \\
& Low-SemSub-Geom ($\alpha$=0.5)  & 51.39 & 96.23 & 66.99 & 52.72 & 93.39 \\
& High-SemSub-Geom ($\alpha$=1.0) & 51.84 & 95.65 & \textbf{67.24} & 53.42 & 92.22 \\
& High-SemSub-Geom ($\alpha$=0.5) & 51.73 & 95.84 & 67.19 & 53.23 & 92.58 \\
\midrule

\multirow{5}{*}{\textit{Adversarial}}
& Low-SemSub-Geom ($\alpha$=1.0)  & \textbf{52.90} & 92.59 & 67.33 & \textbf{54.92} & \textbf{87.82} \\
& Low-SemSub-Geom ($\alpha$=0.5)  & 51.65 & 96.17 & 67.21 & 53.08 & 93.08 \\
& High-SemSub-Geom ($\alpha$=1.0) & 52.22 & 95.76 & \textbf{67.58} & 53.89 & 92.04 \\
& High-SemSub-Geom ($\alpha$=0.5) & 52.03 & 95.91 & 67.46 & 53.71 & 92.24 \\
\bottomrule
\end{tabular}
\label{tab:supp_alpha_ablation_full}
\end{table*}

\section{Additional Ablations}
\label{sec:add-ablation}

\noindent \textbf{Negative Transfer and Query Sensitivity.}
A critical concern in contrastive decoding is whether the ``negative'' context accidentally remains consistent with the query, leading to the suppression of correct answers.
We analyze this trade-off using our geometric noise ablation on the 3D-POPE benchmark in Table~\ref{tab:geom_noise_ablation}.
We observe that minimal distortions ($\varepsilon=0.01$) indeed fail to provide a sufficient contrastive signal, resulting in higher Yes-rates (86.05\%) and lower F1 scores (73.81\%) compared to moderate distortions. This supports the view that if the negative scene is too similar to the original, the penalty is ineffective.
However, performance peaks at moderate noise levels ($\varepsilon=0.05$), achieving the highest F1 (75.00\%) and Accuracy (67.65\%) on the \textit{random} split. Performance degrades again at extreme noise levels ($\varepsilon=0.45$), confirming that structural coherence is necessary for a meaningful contrastive signal.

\noindent \textbf{Sensitivity to Geometric Perturbations.} 
We perform an ablation study to examine how \modelnamenc{} responds to varying levels of geometric noise injected into the 3D scene graph. Specifically, we distort each object’s centroid and extent using zero-mean Gaussian perturbations with increasing standard deviations $\varepsilon \in \{0.01, 0.05, 0.15, 0.45\}$. The goal is to determine whether \modelnamenc{} relies on exact spatial coordinates or whether its contrastive mechanism is robust to structural uncertainty.\looseness-1

In \Cref{tab:geom_noise_ablation}, and across most 3D-POPE splits, moderate distortions yield the highest gains: $\varepsilon = 0.05$ for instance, achieves {F1 = 75.00} and {Accuracy = 67.65} on the \textit{random} split, while reducing Yes-Rate from roughly $86\%$ at $\varepsilon = 0.01$ and $\varepsilon = 0.45$ down to $78.41\%$. A slightly larger perturbation level ($\varepsilon = 0.15$) produces nearly identical performance, suggesting a performance plateau at moderate distortion levels.\looseness-1
Conversely, the smallest and largest distortions ($\varepsilon = 0.01$ and $\varepsilon = 0.45$) degrade performance relative to the mid-range settings. Too little noise appears insufficient to decorrelate hallucination-inducing features, while excessive noise corrupts spatial information necessary for grounding. This pattern mirrors observations in the original 2D VCD literature~\cite{leng2024mitigating}, where blurring an image enough to remove high-frequency appearance features, but not erasing structural content, produces the strongest contrastive signal, whereas both minimal perturbation and extreme degradation lead to diminished gains. Overall, these findings suggest that our method performs best under moderate geometric perturbations that disrupt over-specific coordinate cues while maintaining the high-level spatial layout of the scene. This supports the view that \modelnamenc{} relies on coarse structural grounding rather than fine-grained 3D precision, making it robust to noisy or imperfect scene representations.

\noindent \textbf{Robustness to Perturbation Strength.}
Table~\ref{tab:supp_alpha_ablation_full} presents detailed results across 3D-POPE splits under varying perturbation strengths, combining both semantic and geometric distortions. We evaluate two  $\alpha$ configurations:  $\alpha=1.0$ (strong contrastive signal), and  $\alpha=0.5$ (moderate),
Notably, the Random split yields the strongest performance, achieving an F1 of 74.48\% under \textit{Low-SemSub-Geom} with $\alpha=1.0$. This shows that \modelnamenc{}’s robustness extends to unseen, diverse prompts. 
Overall, increasing $\alpha$ (contrastive strength) produces small but consistent improvements in F1, demonstrating that stronger contrastive supervision amplifies grounding consistency without sacrificing generalization. These results collectively validate that combining semantic and geometric distortions provides the most balanced and stable form of regularization for visual contrastive decoding in 3D embodied environments.

\section{Results on the HEAL Probing Set}
We evaluate the efficacy of our proposed 3D-VCD method on the HEAL benchmark, specifically focusing on adversarial probes designed to induce hallucinations through scene-task inconsistencies (Table~\ref{tab:heal_comparative_results}). 
We apply our method by contrasting the adversarial prompts (\eg, Distractor Injection) against their corresponding clean baseline prompts. This setup tests whether \modelnamenc{} can recover the agent's original grounding capabilities even when presented with misleading textual cues.

Our inference-time intervention significantly enhances the grounding capabilities of the Qwen-14B-Instruct model across all challenging splits. 
Under the \textit{Distractor Injection} setting, which tests resilience to irrelevant text mentions, 3D-VCD reduces the State Hallucination Rate ($C_S$) by approximately 70\% (from 16.5\% to 5.0\%), effectively steering the generation away from ungrounded textual cues and towards the physical scene reality. 
Furthermore, in the \textit{Scene-Object Synonymous} probe, 3D-VCD achieves the lowest Object Hallucination Rate ($C_O$) of 1.0\% among all evaluated models, outperforming even the robust Llama-3-8B baseline. This indicates that our contrastive decoding strategy reinforces semantic consistency, ensuring the agent grounds its reasoning in the object's physical presence rather than its lexical token. 
Most notably, under \textit{Scene-Task Contradiction}, where the base model hallucinates objects 53.9\% of the time to satisfy impossible goals, 3D-VCD drastically mitigates this behavior ($C_O$ 1.5\%), demonstrating a strong refusal to fabricate non-existent items.

\begin{table*}[t]
\centering
\caption{\textbf{Comparative Hallucination Rates (CHAIR) on HEAL Probes.} 
We report Object Hallucination ($C_O$) and State Hallucination ($C_S$) rates.
Our proposed {3D-VCD} method (applied to Qwen-14B) shows reduced hallucination, particularly under Distractor Injection.}
\label{tab:heal_comparative_results}
\vspace{-0.3cm}
\resizebox{\linewidth}{!}{
\begin{tabular}{clcccccccccc}
\toprule
\multirow{3}{*}{\textbf{Env}} & 
\multirow{3}{*}{\textbf{Models}} &
\multicolumn{2}{c}{\multirow{2}{*}{\textbf{Base Prompt}}} &
\multicolumn{4}{c}{\textbf{TaskDescMod}} &
\multicolumn{4}{c}{\textbf{SceneMod}} \\ 
\cmidrule(lr){5-8}\cmidrule(lr){9-12}       
 & & & & 
\multicolumn{2}{c}{\textbf{DistInj}} &
\multicolumn{2}{c}{\textbf{TaskObjRem}} &
\multicolumn{2}{c}{\textbf{SynonymSub}} &
\multicolumn{2}{c}{\textbf{SceneTaskCon}} \\ 
 & & $C_O\downarrow$ & $C_S\downarrow$ &
$C_O\downarrow$ & $C_S\downarrow$ &
$C_O\downarrow$ & $C_S\downarrow$ &
$C_O\downarrow$ & $C_S\downarrow$ &
$C_O\downarrow$ & $C_S\downarrow$ \\
\midrule \midrule

\multirow{5}{*}{\rotatebox[origin=c]{90}{BEHAVIOR}} 
& Gemma-2-9b-it          & 0.87 & 1.80 & 11.60 & 5.70 & 10.8 & 4.8 & 11.4 & 2.8 & 73.1 & 8.2 \\
& DS-R1-Distil-LLaMA-8B  & 1.26 & 8.16 & 13.90 & 27.20 & 6.8 & 13.3 & 20.4 & 10.1 & 60.0 & 19.6 \\
& Llama-3-8B-Instruct    & 1.93 & 2.15 & 2.58 & 9.49 & 5.0 & 6.7 & 2.2 & 7.4 & 17.8 & 7.7 \\
& Qwen-14B-Instruct      & 1.1  & 11.0 & 4.1  & 16.5 & 3.5 & 10.0 & 3.9 & 10.0 & 53.9 & 5.7 \\
\rowcolor{gray!10} 
&  Qwen-14B-Instruct-VCD  & --   & --   & \textbf{3.6} & \textbf{5.00} & 4.7 & \textbf{7.2} & \textbf{1.0} & \textbf{5.7}  & \textbf{1.5} & 7.2 \\
\bottomrule
\end{tabular}
}
\end{table*}

\section{Additional Qualitative Examples}
\begin{figure*} 
    \centering
    \includegraphics[width=0.9\textwidth]{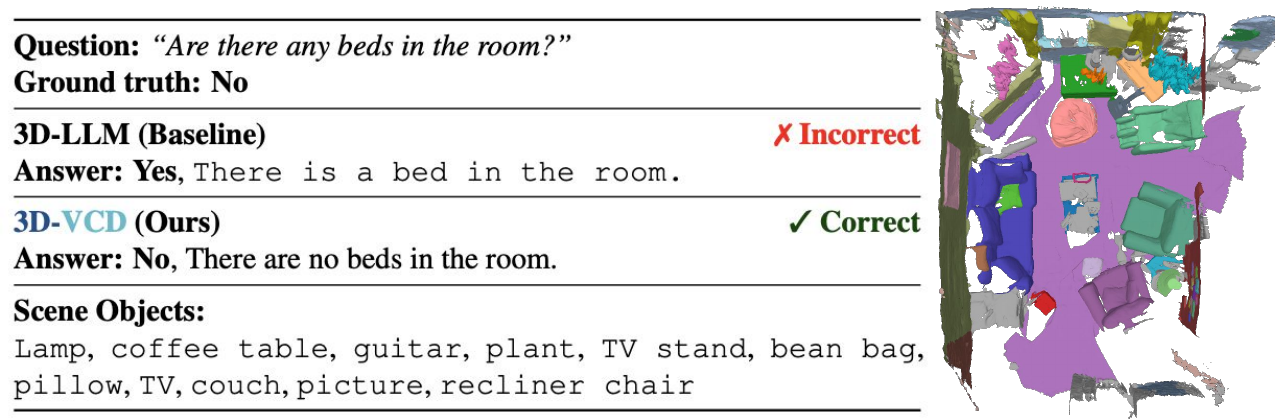}
     \vspace{-0.3cm}
    \caption{\textbf{Qualitative comparison on 3D-POPE.} The baseline 3D-LLM hallucinates the presence of a bed. In contrast, \modelnamenc{} correctly answers \textit{No} by contrasting logits under perturbed 3D scene graphs, effectively suppressing hallucinated object activations. The right panel shows the object-level scene segmentation for reference.}
    \label{fig:beds}
\end{figure*}

\begin{figure*}[t]
    \centering
    \includegraphics[width=0.9\textwidth]{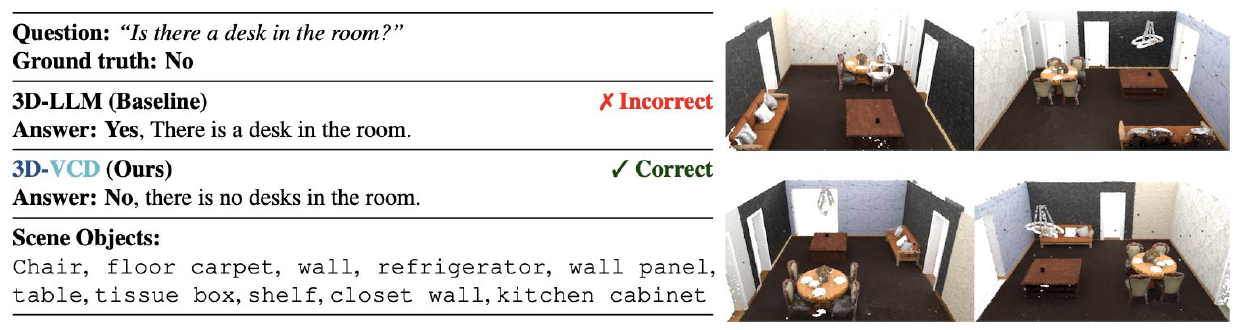}
    \vspace{-0.3cm}
    \caption{
    \textbf{Qualitative comparison on 3D-POPE.} The baseline 3D-LLM hallucinates a desk object and incorrectly predicts its presence. In contrast, \modelnamenc{} correctly determines that no desk exists by suppressing spurious category matches through contrastive decoding aligned with the object-centric scene graph.
    }
    \label{fig:desk}
\end{figure*}

\noindent Across both 3D-POPE and HEAL, our qualitative examples highlight the central deficiencies of current 3D-aware language models. Baseline models often fail at the most fundamental level by producing an incorrect ``yes'' or ``no'' response, as shown in Figures \ref{fig:beds} and \ref{fig:desk}. These examples indicate that the model relies heavily on linguistic priors rather than true 3D evidence, particularly in cluttered scenes or when relational reasoning is involved. 
In the HEAL example (Figure~\ref{fig:beds}), the baseline 3D-LLM hallucinates the presence of a bed and incorrectly answers ``yes,'' despite the scene information indicating otherwise. By contrast, 3D-VCD correctly answers by contrasting logits under the original and perturbed 3D scene graphs, suppressing object activations that are not grounded in the scene. In the 3D-POPE example (Figure~\ref{fig:desk}), the baseline similarly hallucinates a desk and falsely predicts its presence, whereas 3D-VCD correctly determines that no desk exists. 
Together, these qualitative findings demonstrate that \modelnamenc{} strengthens the model's reliance on authentic visual evidence, yielding more faithful and reliable predictions across diverse 3D reasoning tasks.

\section{\modelnamenc{} Example Input and Output}
\begin{figure*}
    \centering
    \includegraphics[width=0.99\linewidth]{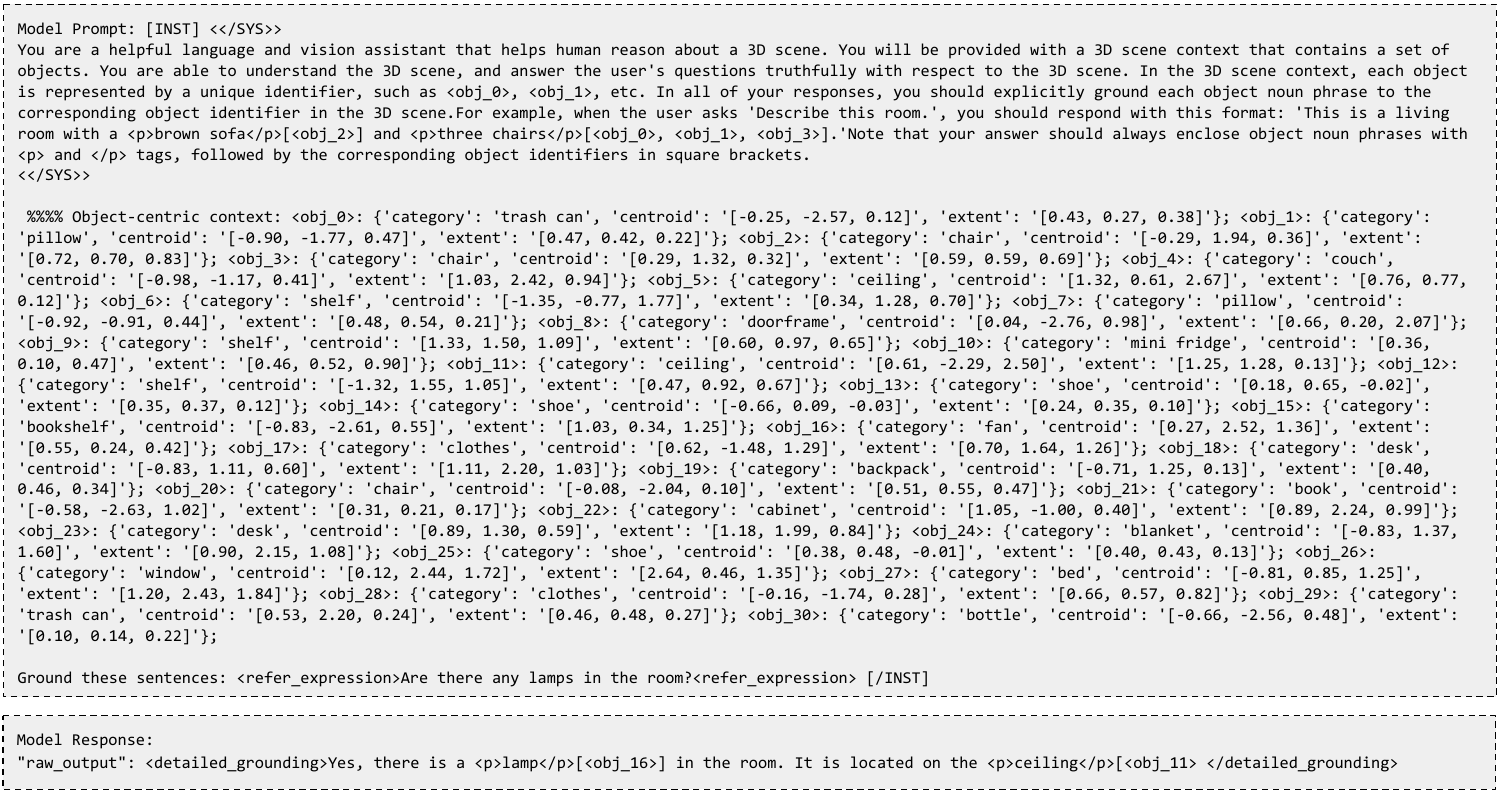}
    \vspace{-0.3cm}
    \caption{\textbf{\modelnamenc{} model input and output.}}
    
    \label{fig:model_input_output}
\end{figure*}

Figure \ref{fig:model_input_output} shows the input and output. For each query, the model must determine whether the mentioned object is present in the 3D scene.
Predictions are parsed from the first grounded response segment enclosed within \texttt{<detailed\_grounding>} tags, \eg,
\texttt{<detailed\_grounding>Yes, there is a <p>chair</p>[<obj\_1>]}.
A response is considered correct only if (1) the binary decision (\texttt{Yes}/\texttt{No}) matches the ground truth,
and (2) the referenced object identifier corresponds to the correct instance in the scene graph.

\section{Broader Impact}
This work aims to improve the reliability of 3D embodied agents by reducing hallucinations at inference time. More trustworthy grounding can benefit applications such as household robotics, assistive systems, and embodied decision-making, where outputs must remain consistent with the observed environment. Because 3D-VCD is training-free and can be applied without retraining, it also offers a practical way to improve existing systems without the cost of collecting new data or updating model parameters.
At the same time, reducing hallucinations does not fully solve reliability in embodied AI. Grounding errors are only one source of failure, and safer perception does not guarantee safe downstream planning or action under noise, distribution shift, or adversarial conditions. As a result, methods such as 3D-VCD should be deployed together with additional safeguards, including uncertainty estimation, human oversight, and rigorous real-world testing.

\end{document}